\documentclass[journal]{IEEEtran}

\usepackage{cite}

\usepackage{multirow}
\usepackage{subfigure}
\usepackage{graphicx}
\usepackage[safe]{tipa}
\usepackage{xcolor}
\usepackage[normalem]{ulem}
\usepackage{amsmath}
\usepackage{amssymb}
\usepackage{caption}
\usepackage[ruled,linesnumbered]{algorithm2e}

\begin{document}

\title{TMMF: Temporal Multi-modal Fusion for Single-Stage Continuous Gesture Recognition}

\author{Harshala~Gammulle,~\IEEEmembership{Member,~IEEE,}
        Simon~Denman,~\IEEEmembership{Member,~IEEE,}
        Sridha~Sridharan,~\IEEEmembership{Life Senior Member,~IEEE,}
        Clinton~Fookes,~\IEEEmembership{Senior Member,~IEEE.}
        \IEEEcompsocitemizethanks{\IEEEcompsocthanksitem H. Gammulle, S. Denman, S. Sridharan and C. Fookes  are with the Signal Processing, Artificial Intelligence and Vision Technologies (SAIVT) Lab, Queensland University of Technology, Brisbane, Australia.\protect\\
E-mail: pranali.gammule@qut.edu.au}
\thanks{This paper has supplementary downloadable material available at https://ieeexplore.ieee.org/Xplore/home.jsp, provided by the author. The material includes additional qualitative and hyper-parameter evaluations. Contact pranali.gammule@qut.edu.au for further questions about this work.}
\thanks{}
\thanks{Manuscript received.}}

\markboth{Journal of \LaTeX\ Class Files,~Vol.~14, No.~8, August~2015}%
{Shell \MakeLowercase{\textit{et al.}}: Bare Demo of IEEEtran.cls for IEEE Journals}

\maketitle

\begin{abstract}
Gesture recognition is a much studied research area which has myriad real-world applications including robotics and human-machine interaction. 
Current gesture recognition methods have focused on recognising isolated gestures, and existing continuous gesture recognition methods are limited to two-stage approaches where independent models are required for detection and classification, with the performance of the latter being constrained by detection performance. In contrast, we introduce a single-stage continuous gesture recognition framework, called Temporal Multi-Modal Fusion (TMMF), that can detect and classify multiple gestures in a video via a single model. This approach learns the natural transitions between gestures and non-gestures without the need for a pre-processing segmentation step to detect individual gestures. To achieve this, we introduce a multi-modal fusion mechanism to support the integration of important information that flows from multi-modal inputs, and is scalable to any number of modes. Additionally, we propose Unimodal Feature Mapping (UFM) and Multi-modal Feature Mapping (MFM) models to map uni-modal features and the fused multi-modal features respectively. To further enhance performance, we propose a mid-point based loss function that encourages smooth alignment between the ground truth and the prediction, helping the model to learn natural gesture transitions.  
We demonstrate the utility of our proposed framework, which can handle variable-length input videos, and outperforms the state-of-the-art on three challenging datasets: EgoGesture, IPN hand and ChaLearn LAP Continuous Gesture Dataset (ConGD). Furthermore, ablation experiments show the importance of different components of the proposed framework.  
\end{abstract}

\begin{IEEEkeywords}
Gesture Recognition, Spatio-temporal Representation Learning, Temporal Convolution Networks.
\end{IEEEkeywords}

\IEEEpeerreviewmaketitle

\section{Introduction}

\IEEEPARstart{G}{estures} play a crucial role in human communication and are used together with a speech to convey meaning. Gestures can be any form of visual action including hand motions, pose changes and facial expressions. Given the vital role of gestures in human communication, automated gesture recognition has been explored in a vast number of application areas including human-computer interaction, robotics, sign language recognition, gaming, and virtual reality control; and due to these diverse applications, automated gesture recognition has received much attention within computer vision research. Gestures, like speech, are continuous in nature, with the next gesture directly related to those that have occurred before. However, despite the temporal relationships that exist within a sequence of gestures, there is limited research that has considered continuous gesture recognition.

Most gesture recognition approaches are based on recognising isolated gestures \cite{joze2020mmtm,molchanov2015hand,molchanov2015multi} where an input video is manually segmented into clips, each of which contains a single isolated gesture. In a real-world scenario where gestures are performed continuously, methods based on isolated gestures are not directly applicable, and thus do not translate to a natural setting. As such, recent approaches \cite{benitez2020ipn,hoang2019continuous,kopuklu2019real} aim to recognise gestures in the continuous original (i.e. unsegmented) video where multiple types of gesture, including both gestures and non-gesture actions, are present. These continuous gesture recognition approaches are formulated in two ways: two-stage \cite{hoang2019continuous,kopuklu2019real,zhu2018continuous} and single-stage \cite{gupta2016online} methods. 

The two-stage approach is built around using two models: one model to perform gesture detection (also known as gesture spotting), and another for gesture classification. In \cite{kopuklu2019real} the authors proposed a two-stage method where gestures are first detected by a shallow 3D-CNN and when a gesture is detected, it activates a deep 3D-CNN classification model. Another work \cite{hoang2019continuous} proposed using a Bidirectional Long Short-Term Memory (Bi-LSTM) to detect gestures while the authors use a combination of two 3D Convolution Neural Networks (3D-CNN) and a Long Short-Term Memory (LSTM) network to process multi-modal inputs for gesture classification. 
\begin{figure}[htbp]
        \centering
        	\includegraphics[width=1.0\linewidth]{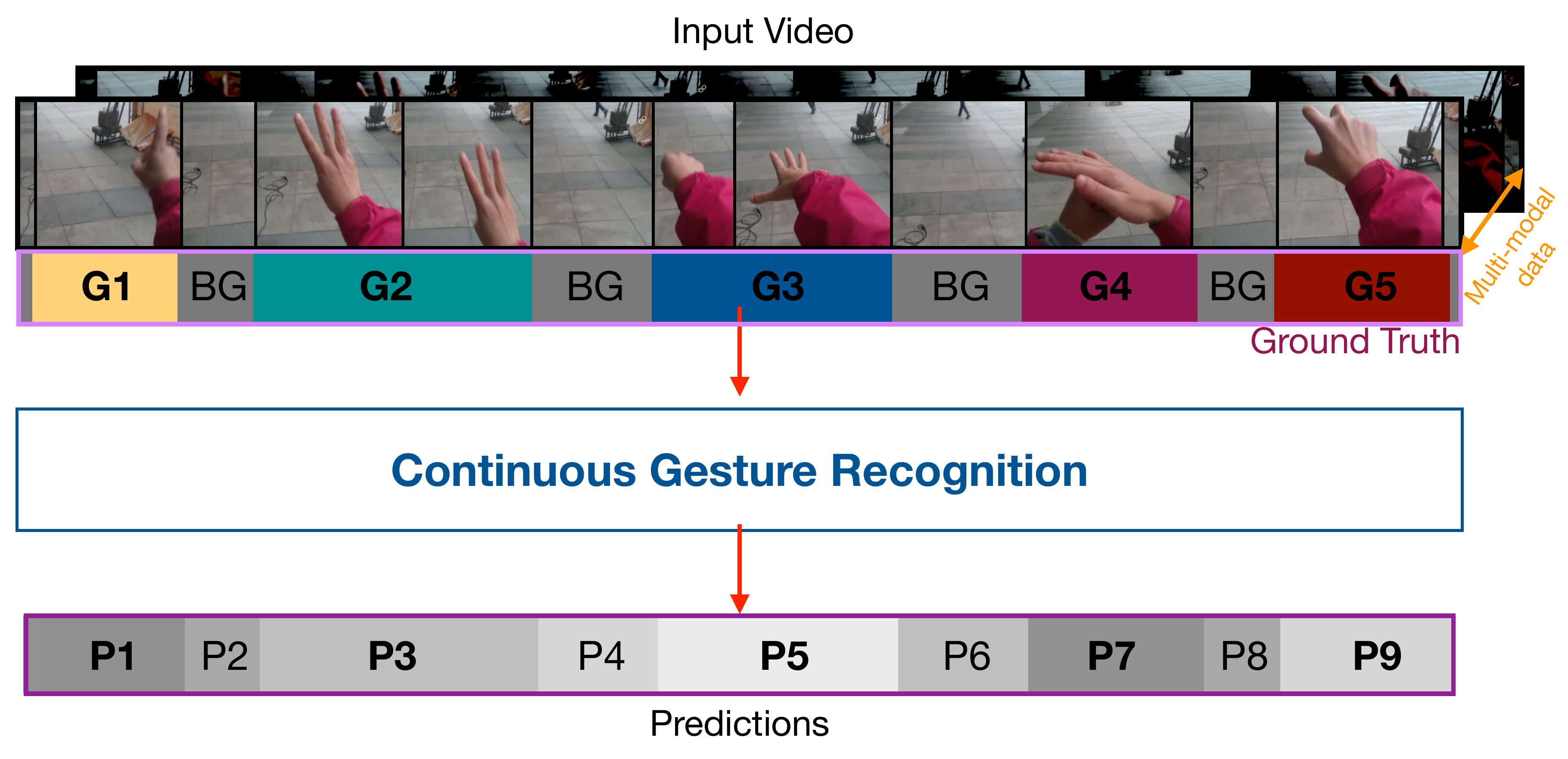}
	\caption{Single-stage Continuous Gesture Recognition: The model is fed with a multi-modal (RGB, Depth etc.) feature sequence and the ground truth label sequence. Input gestures can belong to a particular gesture class or a non-gesture (BG) class. During training, using the ground-truth the model learns to map from the input frames to the correct gesture class.}
	\label{fig:task}
\end{figure}

Single-stage approaches originate from the action recognition domain\cite{lea2017temporal,farha2019ms}, where frames that do not contain an action are labelled `background' (similar to the non-gesture class). In contrast to two-stage methods, single-stage methods use only a single model which directly performs gesture classification. Fig. \ref{fig:task} illustrates the typical structure of a single-stage approach where the recognition is performed by considering all the gesture classes together with the non-gesture class. In addition to being simpler than two-stage methods, single stage methods avoid the potential issue of errors being propagated between stages. For example, in a two-stage method, if the detector makes an error when estimating the start or end of a gesture sequence, this error is propagated through to the classification process. Hence in two-stage methods, classifier performance is highly dependent on the robustness of the detector. However, we observe that two-stage methods are a popular choice among researchers when performing continuous gesture recognition. This is largely due to the challenges that a single network must address when performing both gesture localisation and recognition concurrently.    

Several gesture recognition approaches have also exploited multi-modal data and have shown improved results through fusion \cite{joze2020mmtm,hoang2019continuous}. In \cite{joze2020mmtm}, the authors introduce a simple neural network module to fuse features from two modes for the two-stage gesture recognition task. However, for continuous gesture recognition, any fusion scheme must consider that the input sequence may include multiple gestures that evolve temporally. Hence, using a simple attention layer to fuse domains restricts the learning capacity as model attention is applied to the complete sequence, ignoring the fact that there are multiple gesture sub-sequences, and potentially leading to some individual gestures being suppressed.

In this paper we propose a novel single-stage method for continuous gesture recognition. By using a single-stage approach we expect the classification model to learn natural transitions between gestures and non-gestures. However, directly learning the gestures from a continuous unsegmented video is challenging as it requires the model to detect the transitions between gesture classes, and recognise gestures/non-gestures simultaneously. To improve performance we consider multiple modalities and introduce a novel fusion module that extracts multiple feature sub-sequences from the multi-modal input streams, considering their temporal order. The proposed fusion module preserves this temporal order and enables the learning of discriminative feature vectors from the available modalities. To aid model learning, we propose a novel mid-point based loss function, and perform additional experiments to demonstrate the effectiveness of the proposed loss.

\begin{figure*}[ht]
        \centering
        	\includegraphics[width=0.75\textwidth]{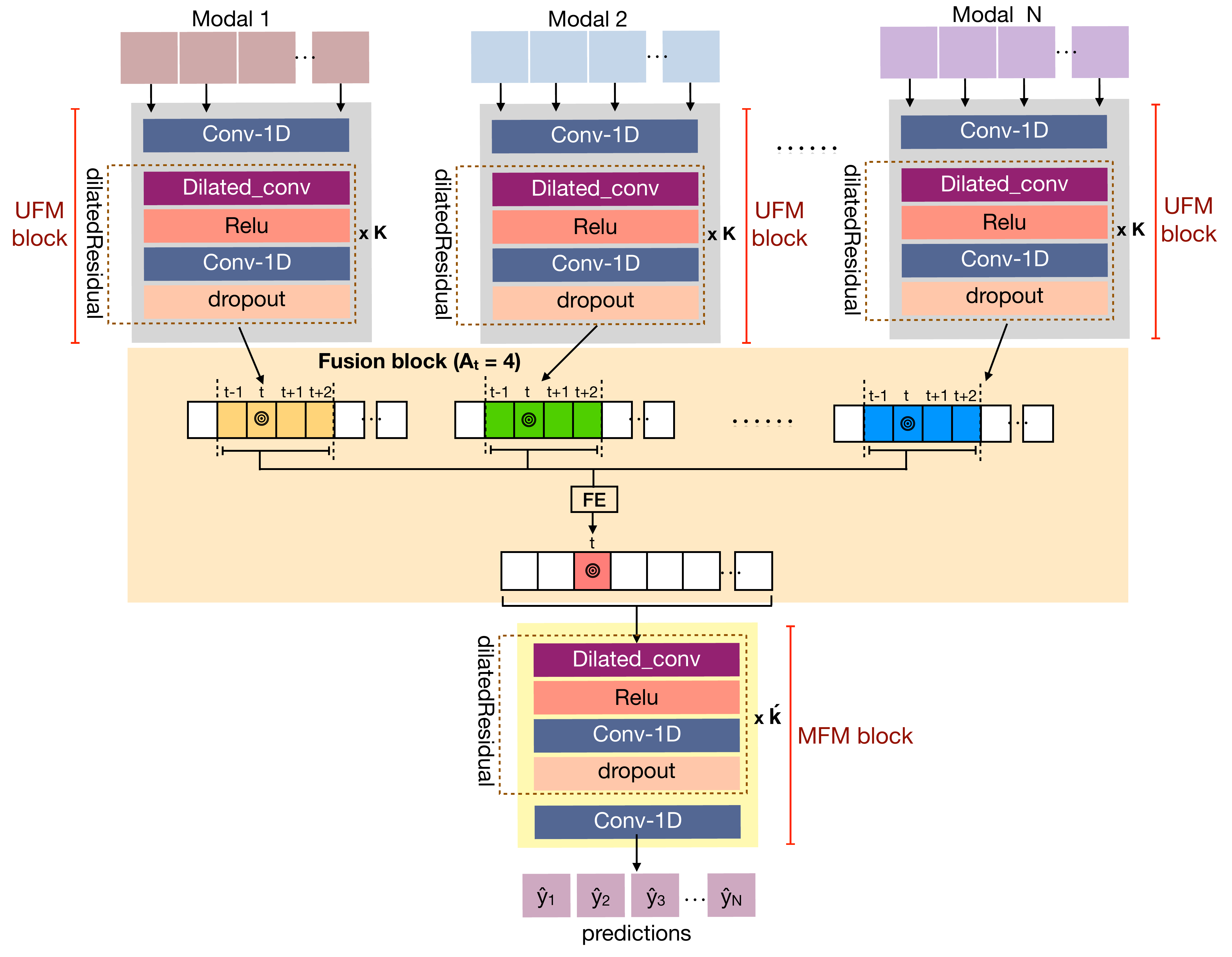}
	\caption{Proposed single-stage Temporal Multi-Modal Fusion (TMMF) framework: The data from each mode is passed through a pre-trained feature extractor and subsequently through separate Unimodal Feature Mapping (UFM) blocks. The output of each UFM block is fused by the proposed fusion block which learns discriminative features from each mode, considering the temporal order of the data. This aids  gesture classification which is performed by the Multi-modal Feature Mapping (MFM) block.}
	\label{fig:model}
\end{figure*}

Figure \ref{fig:model} illustrates the architecture of our proposed Temporal Multi-Model Fusion (TMMF) framework. In the first stage of the model, semantic features from each mode are extracted via a feature extractor, and are passed through a Unimodal Feature Mapping (UFM) block. We maintain separate UFM blocks for each stream. The outputs of all UFM blocks are used by the proposed fusion module which learns multi-modal spatio-temporal relationships to support recognition. The output of the fusion model is passed through the Multi-modal Feature Mapping (MFM) block which performs the final classification. The model is explicitly designed to handle variable length video sequences. Through evaluations on three continuous gesture datasets, ChaLearn LAP Continuous Gesture Dataset (ConGD) \cite{wan2016chalearn}, EgoGesture \cite{zhang2018egogesture} and IPN hands \cite{benitez2020ipn}, we show that our proposed method achieves state-of-the-art results. We also perform extensive ablation evaluations, demonstrating the effectiveness of each component of the model. Furthermore, we illustrate the scalability of our proposed TMMF model by performing continuous gesture recognition with two and three modalities, where the third modality is obtained using the Generative model in \cite{isola2017image}. 

In summary our contributions are three fold:
\begin{itemize}
    \item We propose a novel single-stage temporal multi-modal fusion (TMMF) framework, that uses a single model to recognise gestures and their temporal transitions through multi-modal feature fusion, supported by our proposed fusion block algorithm.
    \item Our model automatically learns the natural transitions between gestures through the proposed mid-point-based loss function.
    \item We carry out experiments showing the model's ability to outperform the state-of-the-art on three challenging datasets, and ablation experiments emphasise the contribution of each component of the proposed TMMF architecture.
\end{itemize}


\section{Related Works}


Gestures are primarily the movements of the limbs or body that aid and emphasize speech, and they play a major role in communication between humans, and in human-computer interactions. Therefore, gesture recognition has been an extensively studied area in computer vision as it enables numerous application areas that require human-computer collaboration. We discuss the related works in the areas of isoloated gesture recognition (see Section \ref{subsec:litiso}), continuous gesture recognition (see Section \ref{subsec:litcont}), and multi-modal gesture recognition (see Section \ref{subsec:litmulti}).

\subsection{Isolated Gesture Recognition}
\label{subsec:litiso}
Isolated gesture recognition uses segmented videos containing a single gesture per video, and is a naive and simplified way of performing gesture recognition which does not reflect the real-world challenge that the gesture recognition presents.

Early discrete gesture recognition methods are based on handcrafted features \cite{wan2015explore,shen2012dynamic,trinh2012hand,yang2014super}. For example, in \cite{wan2015explore} the authors proposed a spatio-temporal feature named Mixed Features around Sparse key-points (MFSK), which is extracted from RGB-D data. In \cite{shen2012dynamic} the authors propose to extract a visual representation for hand motions using motion divergence fields. Other methods are based on extracting Random Occupancy Pattern (ROP) features \cite{wang2012robust}, Super Normal Vectors (SNV) \cite{yang2014super}, and improved dense trajectories \cite{wang2013action}. However, these hand-crafted feature methods rely on human domain knowledge and risk failing to capture necessary information that may greatly contribute towards correct recognition. 

Subsequently, attention has shifted to deep network-based approaches \cite{simonyan2014two,gammulle2019predicting,teng2019deep,shou2016temporal,ji20123d} due to their ability to learn task-specific features automatically. As such, most recent gesture recognition methods use deep networks \cite{zhang2019eleatt,liu2017continuous,kopuklu2019real,benitez2020ipn,joze2020mmtm,li2020one,su2017unsupervised} and have demonstrated superior results to their hand-crafted counterparts. 

In \cite{zhang2020gesture}, the authors proposed three variants of 3D CNNs which are able to learn spatio-temporal information through their hierarchical structure to learn to recognise isolated gestures. The authors of \cite{joze2020mmtm} proposed a fusion unit to integrate and learn information that flows through two uni-modal CNN models to support isolated gesture recognition. \cite{abavisani2019improving} introduced a multi-modal training/uni-modal testing approach where the authors embed the knowledge from individual 3D-CNN networks, forcing them to collaborate and learn a common semantic representation to recognise isolated gestures. However, the simplicity of isolated gesture recognition methods prevents their direct real-world application, as input videos that contain multiple gestures must first be segmented. To address this, \cite{benitez2021improving} conducted experiments using light-weight semantic segmentation methods such that an isolated gesture based method could be used in real-time. An alternate approach is the development of methods to directly recognise gestures in unsegmented (i.e. continuous) video streams \cite{kopuklu2019real,liu2017continuous}.  


\subsection{Continuous Gesture Recognition}
\label{subsec:litcont}
Continuous Gesture Recognition is evaluated on unsegmented gesture videos, and contains more than one gesture per video.  

In such unsegmented videos, as there are sub-sequences containing both gestures and non-gestures, a typical model first detects gesture regions (also known as gesture spotting \cite{zhang2018egogesture}) prior to recognising each gesture. \cite{kopuklu2019real} formulated a two-stage framework to carry out the detection and classification of continuous gestures, where their detection method first performs gesture detection and the classification model is activated only when a gesture is detected. However, these two-stage methods require two separate networks to perform gesture detection and classification respectively. This limitation has motivated us to develop a single-stage method which requires only a single model to learn the gestures and their natural transitions.

To the best of our knowledge \cite{gupta2016online, zhang2018egogesture} are the only existing single-stage continuous gesture recognition methods. In \cite{gupta2016online} the authors employ an RNN to predict the gesture labels for an input frame sequence. In \cite{cao2017egocentric} the authors use a C3D model to classify continuous gestures. The model sequentially slides over the input video and outputs a single gesture class representing the gesture within that input window, including the non-gesture class. They propose to further improve the gesture prediction method by employing a Spatio-Temporal Transfer Module (STTM) \cite{cao2017egocentric} and an LSTM network, where the LSTM predicts gesture labels based on the C3D features. However, these methods fail to achieve the accuracy of two-stage methods. We believe this is due to the simplistic nature of the architecture, which cannot not handle the complexities of the single-stage formulation.

The related task of continuous action recognition (also known as temporal action segmentation) has been approached using various strategies \cite{lea2017temporal,farha2019ms,gammulle2019coupled,gammulle2020fine}. Unlike gesture recognition, most temporal action segmentation approaches are single-stage methods where detection and classification is performed by a single network. Single-stage methods offer advantages over two-stage methods in that there is only a single model and errors from the first stage (the detector) are not propagated to the second stage (the classifier). Furthermore, a single stage model can learn not only a single gesture sequence, but also leverage information on how different types of gestures are sequentially related. \cite{lea2017temporal} introduced Temporal Convolution Networks (TCN) that use a hierarchy of temporal convolutions. In \cite{farha2019ms}, the authors have extended the ideas of \cite{lea2017temporal} and introduced a multi-stage model for action segmentation, where each stage is composed of a set of dilated temporal convolutions that generate predictions at each stage. We take inspiration from \cite{lea2017temporal, farha2019ms} and make use of temporal convolutions and residual dilated temporal convolutions when formulating our uni-modal and multi-model feature mapping blocks (i.e. UFM and MFM blocks). However, the models in \cite{lea2017temporal, farha2019ms} are unsuitable for a multi-modal problem. Hence we design our gesture recognition model to exploit multi-modal data. 

\subsection{Multi-modal Gesture Recognition}
\label{subsec:litmulti}

Multi-model methods have been investigated in multiple research areas \cite{vielzeuf2018centralnet,Reviewer1_sugg1,simonyan2014two}. These methods can either use multi-modal fusion (early, intermediate or late fusion) \cite{simonyan2014two,owens2018audio}, or learn uni-modal networks through multi-modal network training \cite{abavisani2019improving}.

In \cite{chen2017multimodal}, an early fusion hard-gated approach is proposed for multi-modal sentiment analysis, while in \cite{simonyan2014two} a late fusion method (fusion at the prediction level) is proposed for action recognition. In the gesture recognition domain, the authors in \cite{joze2020mmtm} proposed a simple neural network unit to fuse information from the intermediate features of uni-modal networks. Their proposed unit can be added at different levels of the feature hierarchy with different spatial feature dimensions. In \cite{abavisani2019improving}, the authors utilise separate networks for each available modality and encourage the networks to share information and learn common semantic features and representations to improve the individual networks, and achieve better performance. However, the methods in \cite{joze2020mmtm, abavisani2019improving} have limited applicability to the single stage recognition paradigm and are designed only to handle segmented actions or gestures. When sequences are segmented all frames are part of the same gesture, hence, a simple attention or concatenation of features can produce good results as all the information relates to the one gesture. In contrast, in a single-stage model operating over continuous gestures, the input to the classifier contains multiple gesture sequences and non-gesture frames. Hence, the fusion strategy should understand how these sub-sequences are temporally related and filter the most relevant information considering this temporal order. To this end, we introduce a fusion mechanism that preserves this temporal accordance, and that can be applied to two or more modalities for continuous gesture recognition. 

In \cite{Reviewer1_sugg1,Reviewer1_sugg2} the authors proposed multi-modal methods which share information through rich labels from the text domain in order to handle the problem of insufficient image training data through the proposed Deep Transfer Networks (DTN). Both models take a similar approach to ours where the uni-modal information (text-domain and image domain) is first learned through separate uni-modal networks, and they learn a shared intermediate representation. However, in our proposed method we learn from both spatial and temporal information, and explicitly learn the temporal evolution of the spatial information across multiple modalities.    

\section{Method}


We introduce a novel framework, temporal multi-modal fusion (TMMF) model, to support multi-modal single-stage video-based classification of gestures. In the introduced framework, first videos from each mode are passed through feature extractors, and the extracted deep features are subsequently passed through individual unimodal networks which we term Unimodal Feature Mapping (UFM) blocks. The output feature vector of each UFM block is used by the proposed fusion block to create a discriminative feature vector, which is passed to the Multi-modal Feature Mapping (MFM) block to perform classification. Figure \ref{fig:model} illustrates the overall model architecture.               

The task our approach seeks to solve can be defined as follows: given a sequence of video frames $X^i = \{x^i_{1}, x^i_{2},\dots, x^i_{T}\}$, where $i= 1, 2, \dots, M$ (M is the number of modalities), we aim to infer the gesture class label for each time step $t$ (i.e $ \hat{y}_{1}, \hat{y}_{2}, \dots,\hat{y}_{T}$). 

Our TMMF framework can be used with segmented or unsegmented videos, that may be composed of one or more gesture classes, and supports the fusion of any number of modalities greater than 1. Each feature stream has it's own UFM block that is used to obtain a domain specific representation prior to fusion. In the following sections, we provide a detailed description of the models and the proposed loss formulation.    

\subsection{Unimodal Feature Mapping (UFM) Block}

Video frames for a given modality are passed through a feature extractor (each mode has it's own feature extractor to learn a mode specific representation), and the extracted features are the input to the UFM block. Through the UFM block, we capture salient features related to a specific modality and learn a feature vector suitable for feature fusion. As shown in Figure \ref{fig:model}, this uni-modal network is composed of temporal convolution layers and multiple dilated residual blocks, where each dilated residual block is composed of a dilated convolution layer followed by a non-linear activation function and a $1\times1$ convolution-BatchNorm-ReLu \cite{isola2017image} layer. A dilated convolution can be defined as a convolution where the filter is applied over an area larger than its length by skipping input values at a defined interval. It is similar to a convolution with a larger filter where zeros are placed within the filter to achieve the dilation effect, however dilated convolutions are considered more efficient \cite{oord2016wavenet}.

We take inspiration from \cite{farha2019ms}, where the authors use residual connections to facilitate gradient flow. As in \cite{oord2016wavenet,farha2019ms,lea2017temporal}, we use a dilation factor that is doubled at each layer and each layer is composed of an equal number of convolution filters. Our UFM block has a similar architecture to the single-stage model utilised in \cite{farha2019ms}, however without the final prediction layer. In \cite{farha2019ms}, multiple single-stage models are stacked together in order to formulate the multi-stage architecture, while at each single-stage an action prediction is made which is then refined by the next stage. It should be noted that we adopt the UFM block only to encode the features (no gesture predictions are made before feature fusion) and learn the temporal uni-modal data in a manner that supports the feature fusion and gesture segmentation tasks. We further illustrate the importance of the UFM block through ablation experiments in Section \ref{subsec:ablation_exp}.

\subsection{Fusion Block}

\begin{figure}[htbp]
        \centering
        	\includegraphics[width=0.9\linewidth]{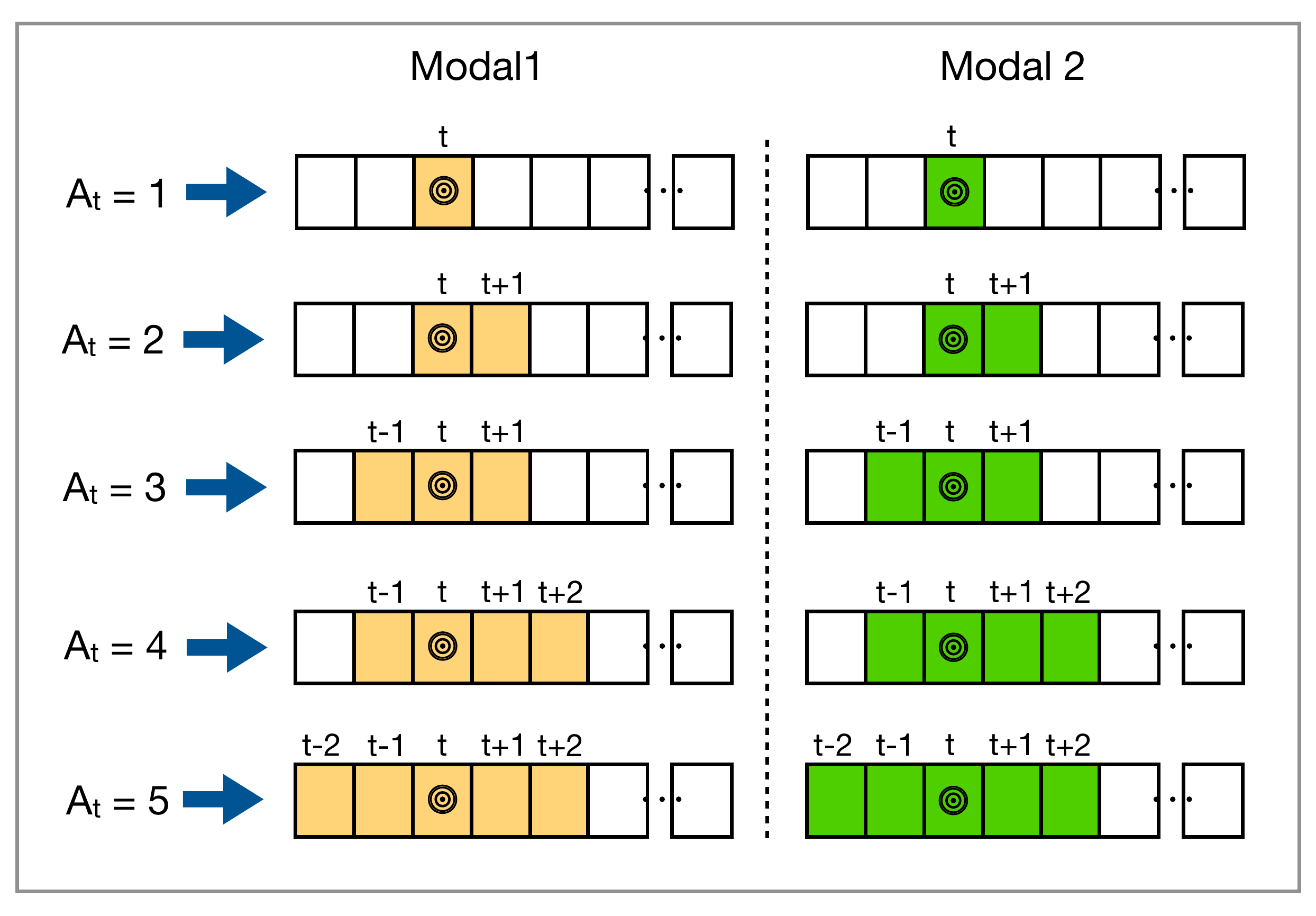}
	\caption{Illustration of the proposed Attention level parameter, $A$, and the associated attention scheme. This parameter determines the number of temporal neighbours that a particular frame is associated with, controlling the information flow to the fusion module. For instance if $A=5$ two neighbouring feature units surrounding the current time step $t$ in each direction (i.e from t-2 to t and from t to t+2) are selected and processed.}
	\label{fig:attentions}
\end{figure}

The output vectors of the UFM blocks are passed through the fusion block, which extracts temporal features from the uni-modal sequences, considering their temporal accordance with the current time step. Feature fusion is performed using the attention level parameter. This parameter defines the feature units that should be selected from the output vector of each UFM block at a given time. An illustration is given in Figure \ref{fig:attentions}.  

\subsubsection{Attention Level parameter ($A$)} Let $V^1_t, V^2_t, \dots, V^M_t$ be the output feature vectors from the UFM blocks representing the $M$ modalities, where $t = {1,2,\dots, T}$. By considering the value set for the parameter $A$ the algorithm decides which feature units from each vector should be selected for fusion at time $t$. This selection criteria is defined based on the fact that multi-modal feature streams are synchronised and the features from the temporal neighbours at a particular timestamp should carry knowledge informative for the gesture class of that frame, while distant temporal neighbours do not carry helpful information (as they are likely from different gesture classes). Based on whether $A$ is even or odd, we calculate the position increment ($i_{inc}$) and decrement ($i_{dec}$) values as shown below. Here, $i_{inc}$ defines the number of units ahead we should consider during the fusion, while $i_{dec}$ defines the number of units behind that should be selected.

\begin{algorithm}[htb]
\SetAlgoLined
\KwIn{$A$: Attention Level}
\KwOut{ $i_{inc}$ and $i_{dec}$}
\uIf{$A$ is even (i.e. $A\%2 = 0$)}{
    $i_{inc} = A/2$ and $i_{dec} = (A-2)/2$ \;
  }
  \uElseIf{$A_{t}$ is odd (i.e. $A\%2 = 1$)}{
    $i_{inc} = i_{dec} = (A-1)/2$
  }
\Return $i_{inc}$, $i_{dec}$ 
\caption{Calculation of position increment ($i_{inc}$) and decrement ($i_{dec}$) values based on the Attention level parameter.}
\label{alg:alg1}
\end{algorithm}

\begin{figure}[htbp]
        \centering
        	\includegraphics[width=0.8\linewidth]{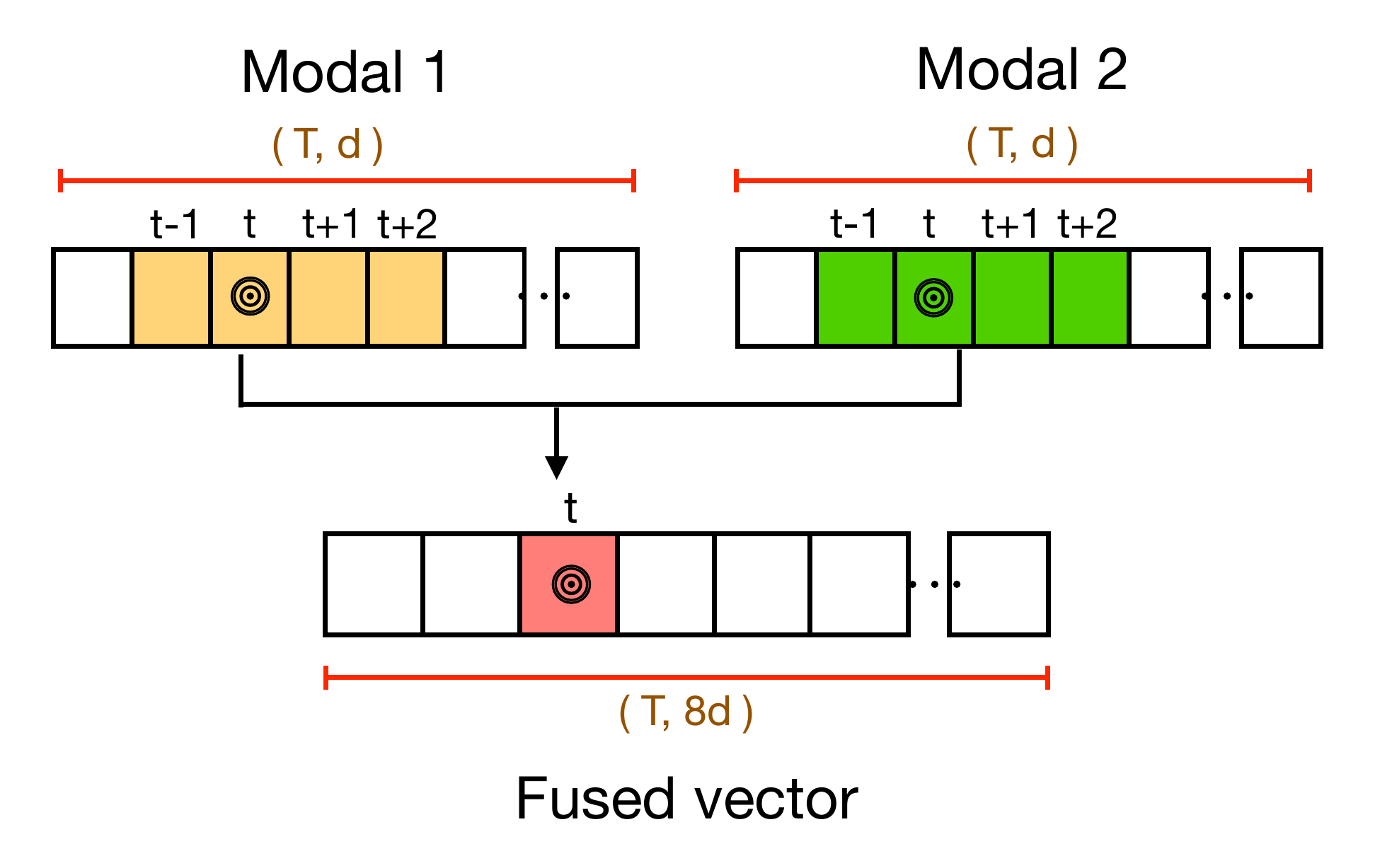}
	\caption{Illustration of the fusion process when $A = 4$. Features surrounding the current time step $t$ are passed to the proposed fusion block from each modality and it first concatenates them (See. Sec. \ref{sec:fe}). Then it passes the concatenated feature vector through the proposed feature enhancement function which identifies salient feature values in the concatenated vector to enhance, and components to suppress. Utilising this scheme we identify the most informative feature units from the local temporal window for decision making at the current time step.}
	\label{fig:fusion}
\end{figure}

Once $i_{inc}$ and $i_{dec}$ are calculated, at $t$ the units from $t-i_{dec}$ to $t+i_{inc}$ are selected from each feature vector. This sub feature vector is given by,
\begin{equation}
    S^i_t = [V^i_{t-i_{dec}}, \dots, V^i_t, \dots, V^i_{t+i_{inc}}],
\end{equation}
where $i = 1, 2, \dots, M$. As shown in Figure \ref{fig:fusion}, when the attention level is 4, 4 feature units (from $t-1$ to $t+2$) are selected from each vector from the UFM. 

\subsubsection{Feature Enhancer (FE)}
\label{sec:fe}
At each time step $t$, the feature enhancer receives computed sub-vectors $S^i_t$s from each UFM, where $i$ indicates the modality, and concatenates these sub-vectors generating an augmented vector $\eta_t$,
\begin{equation}
    \eta_t = [S^1_t, \ldots, S^i_t, \ldots, S^M_t].
\end{equation}
If each feature unit is of dimension $d$ and the attention level is $A$, then $\eta_t$ will have shape, ($d, A \times M$). We then utilise the proposed Feature Enhancer (FE) block, which is inspired by the squeeze and excitation block architecture introduced in \cite{hu2019squeeze}, to allow the model to identify informative features from the fused multi-modal features, enhancing relevant feature units and suppressing uninformative features. However, the squeeze-and-excitation block of \cite{hu2019squeeze} considers the overall 2D/3D CNN layer output and enhances features considering their distribution across channels. In contrast, we propose to enhance features within sub-feature vectors, $V^i$s, for each $t$. Through the FE block, features from each sub-feature vector are enhanced by explicitly modelling the inter-dependencies between channels, further supporting the multi-modal fusion. To exploit the sub-feature dependencies we first perform global average pooling to retrieve relevant information within each of the $d$ channels of the sub-feature vector. This can be defined by,
\begin{equation}
    z_t = F^{GAP}(\eta_t(a,m)) = \frac{1}{ A \times M}\sum_{a=1}^{A}\sum_{m=1}^{M} \eta_t(a,m), 
\end{equation}
Then a gating mechanism implemented using sigmoid activations is applied to filter out the informative features within $d$ such that,
\begin{equation}
    \beta_t = \sigma(W^2\times \mathrm{ReLu}(W^1 \times z_t)),
\end{equation}
where $W^1$ and $W^2$ are trainable weights of the gating mechanism. An augmented feature vector, $\tilde{\eta} $, is obtained by multiplying the sub feature vector $\eta_t$ by the respective weights. This feature vector is also of shape ($d, A \times M$), however, the informative information within is enhanced by considering all modalities.

\subsection{Multi-modal Feature Mapping (MFM) block} 

The MFM block learns to generate the final gesture classification for frame $t$ using the fused feature vector $\tilde{\eta_t}$. Similar to the UFM, the MFM uses a series of temporal convolution layers and multiple dilated residual blocks to operate over the fused feature vector, $\tilde{\eta} = [\tilde{\eta_1}, \ldots, \tilde{\eta_T}]$. By considering sequential relationships, it generates the frame-wise gesture classifications, $\hat{y}_{1}, \dots,\hat{y}_{T}$. This can be written as,
\begin{equation}
    \hat{y}_{1}, \dots,\hat{y}_{T} = F^{\mathrm{MFM}} ([ \tilde{\eta_1}, \ldots, \tilde{\eta_T}]).
\end{equation}

\subsection{Loss Formulation} 
As the classification loss we utilise the cross-entropy loss which is defined as,
\begin{equation}
    \mathcal{L}_{ce} = \dfrac{1}{T} \sum_{t} -\log(\hat{y}_t). 
\label{eq:ce}
\end{equation}

However, only using the frame wise classification loss to learn gesture segmentation is insufficient and can lead to over segmentation errors, even while maintaining high frame wise accuracy. Hence, we also use the smoothing loss introduced by \cite{farha2019ms}. This smoothing loss uses the truncated mean squared error over the frame-wise log probabilities. The smoothing loss can be defined as,
\begin{equation}
    \mathcal{L}_{sm} = \dfrac{1}{T \times C} \sum_{t,c} \tilde{\Delta}_{t,c},
\label{eq:sm}
\end{equation}
where,
\begin{equation}
    \tilde{\Delta}_{t,c}= 
    \begin{cases}
        \Delta_{t,c},& \text{if } \Delta_{t,c}\leq \tau\\
        \tau,              & \text{otherwise}
    \end{cases}
\end{equation}
and, 
\begin{equation}
    \Delta_{t,c} = \lvert \log \hat{y}_{t,c} - \log \hat{y}_{t-1,c} \rvert.
\end{equation}

Here, $T$, $c$, $y_{t,c}$ define the number of frames per sequence, number of classes and the probability of class $c$ at time $t$, respectively. 


In addition to the above two loss functions we further support the action segmentation task through our proposed mid-point loss. However, instead of merely smoothing the predictions, we calculate the distance between the smoothed ground truth and predictions, incorporating a smoothing effect when calculating the loss. Through the mid-point loss, we seek to avoid calculating the loss based on all frames in the sequence, and thus limit the complexity of loss calculations to reduce the overhead that confusion that can occur during the learning process. Using this loss the model is able to capture the dominant classes within the input sequence and pay attention to those classes without been overwhelmed by the small fine-grained details within the windows.   

Let $w$ represent a sliding window with $N$ elements. First, we obtain the ground truth gesture class at the mid-point within the window $w$,
\begin{equation}
    \bar{y} = F^{\mathrm{mid-point}}(y_n) ,
\end{equation}
where $n \in N$. Similarly we obtain the predicted gesture class at the mid-point using,
\begin{equation}
    \tilde{y} = F^{\mathrm{mid-point}}(\hat{y}_n).
\end{equation}

Then we define our mid-point smoothing loss,
\begin{equation}
    \mathcal{L}_{mid} = \sum_{T}\lVert \bar{y} - \tilde{y} \rVert ^2.
\label{eq:med}
\end{equation}

As we are operating over the smoothed ground truth and predicted sequences instead of raw sequences, we observe that this loss accounts for smooth alignment between ground truth and predictions.

Finally, all three loss functions are summed to form the final loss,
\begin{equation}
    \mathcal{L} = \mathcal{L}_{ce} + \lambda_1 \mathcal{L}_{sm} + \lambda_2 \mathcal{L}_{mid},
\label{eq:overall}
\end{equation}
where, $\lambda_1$ and $\lambda_2$ are model hyper-parameters that determine the contribution of the different losses.

\subsection{Implementation details}
The features are extracted from the flatten layer of ResNet50 \cite{resnet}. Prior to the feature extraction, ResNet50 is initialised with weights pre-trained on ImageNet \cite{russakovsky2015imagenet}, and fine-tuned by keeping the first 6 layers frozen. The UFM block is composed of $k=12$ dilated residual layers, while the MFM block contains $k'=10$ dilated residual layers. Similar to \cite{farha2019ms}, we double the dilation factor at each layer. We use the Adam optimiser with a learning rate of 0.0005. The implementation of the proposed framework is completed using PyTorch \cite{paszke2019pytorch}. 

\section{Experiments}

\subsection{Datasets} We evaluate our TMMF model on three challenging public datasets: EgoGesture \cite{zhang2018egogesture}, IPN hand \cite{benitez2020ipn} and the ChaLearn LAP Continuous Gesture dataset \cite{wan2016chalearn}. All three datasets are comprised of unsegmented videos containing one or more gesture per video.

\textbf{EgoGesture dataset \cite{zhang2018egogesture}} is the largest egocentric gesture dataset available for segmented and unsegmented (continuous) gesture classification and is composed of static and dynamic gestures. The dataset contains various challenging scenarios including a static subject with a dynamic background, a walking subject with a dynamic background, cluttered backgrounds, and subjects facing strong sunlight. In our work we utilise the unsegmented continuous gesture data which is more challenging as it requires us to segment and recognise the gesture in a single pass. The dataset consists of 84 classes (83 gesture classes and the non-gesture class) recorded in 6 diverse indoor and outdoor settings. The dataset contains 1,239, 411 frames and 431 videos for training, validation and testing purposes respectively. The dataset provides RGB and depth videos. 

\textbf{IPN hand dataset \cite{benitez2020ipn}} is a recently released dataset that supports continuous gesture recognition. The dataset contains videos based on 13 static/dynamic gesture classes and a non-gesture class. The gestures are performed by 50 distinct subjects in 28 diverse scenes. The videos are collected under extreme illumination conditions, and static and dynamic backgrounds. The dataset contains a total of 4,218 gesture instances and 800,491 RGB frames. Compared to other publicly available hand gesture datasets, IPN Hand includes the largest number of continuous gestures per video, and has the most rapid transitions between gestures \cite{benitez2020ipn}. We utilise the IPN hand dataset specifically as it is provided with multiple modalities: RGB, optical flow and hand segmentation data; which enables us to demonstrate the scalability (Sec. \ref{scalability}) of the proposed framework. 

\textbf{ChaLearn LAP Continuous Gesture Dataset (ConGD) \cite{wan2016chalearn}} is large-scale gesture dataset derived from the ChaLearn Gesture Dataset (CGD). The dataset includes 22,535 RGB-D gesture videos containing 47,933 gesture instances, and each video may contain one or more gestures. Overall the dataset is composed of 249 different gesture classes performed by 21 different subjects.

\subsection{Evaluation Metrics}
\label{metrics}
\textbf{Mean Jaccard Index (MJI)}: To enable state-of-the-art comparisons, we utilise the MJI to evaluate the model on the EgoGesture dataset as suggested in \cite{zhang2018egogesture,wan2016chalearn}. For a given input, the Jaccard index measures the average relative overlap between the ground truth and the predicted class label sequence. The Jaccard index for the $i^th$ class is calculated using,
\begin{equation}
    J_{s,i} = \dfrac{G_{s,i} \cap P_{s,i}}{G_{s,i} \cup P_{s,i}},
\end{equation}
where $G_{s,i}$ and $P_{s,i}$ represents the ground truth and predictions of the $i^{th}$ class label for sequence $s$ respectively. Then the Jaccard index for the sequence can be computed by,
\begin{equation}
    J_s = \dfrac{1}{l_s} \sum_{i=1}^L J_{s,i},
\end{equation}
where $L$ is the number of available gesture classes and $l_s$ represents unique true labels. Then, the final mean Jaccard index of all testing sequences is calculated,
\begin{equation}
    \bar{J_s} = \dfrac{1}{q} \sum_{j=1}^q J_{s,j}. 
\end{equation}

\textbf{Levenshtein Accuracy (LA)}: In order to evaluate the IPN hand dataset we use the Levenshtein accuracy metric used by  \cite{benitez2020ipn}. The Levenshtein accuracy is calculated by estimating the Levenshtein distance between the ground truth and the predicted sequences. The Levenshtein distance counts the number of item-level changes between the sequences and transforms one sequence to the other. However, after obtaining the Levenshtein distances, the Levenshtein accuracy is calculated by averaging this distance over the number of true target classes ($T_p$), and subtracting the average value from 1 (to obtain the closeness), and multiplied by 100 to obtain a percentage,
\begin{equation}
    LA = 1 - \dfrac{l_d}{T_p} \times 100 \%.
\end{equation}

\subsection{Evaluations}
\subsubsection{Selection of Parameter Value, $A$}

\begin{figure*}
    \centering
    \subfigure[EgoGesture]{\includegraphics[width = 0.3\textwidth]{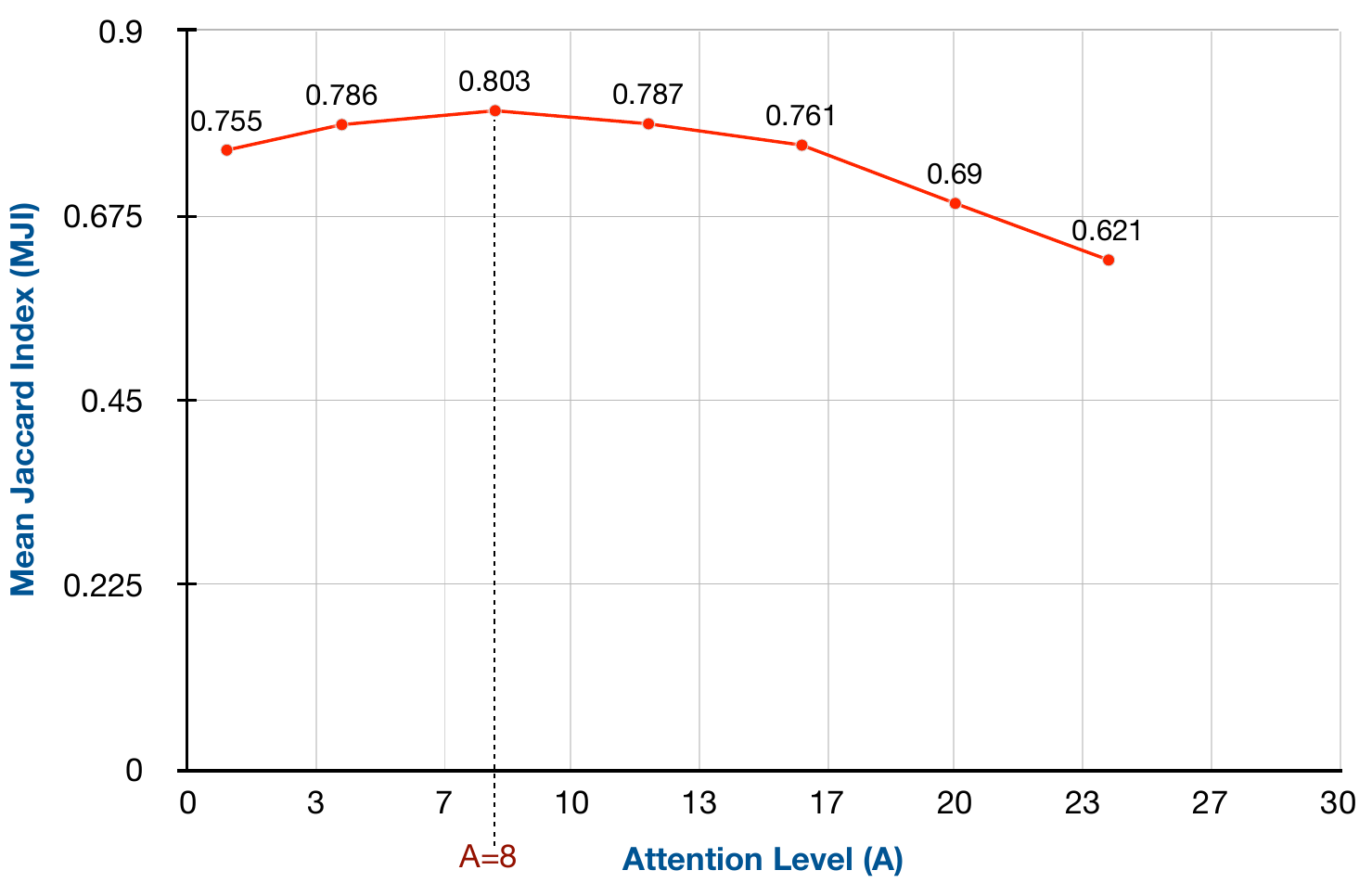}
    \label{fig:attention_level_ego}}
    \subfigure[IPN Hand]{\includegraphics[width = 0.3\textwidth]{Figures/att_graph.pdf}
    \label{fig:attention_level_ipn}}
    \subfigure[ConGD]{\includegraphics[width = 0.3\textwidth]{Figures/att_graph.pdf}
    \label{fig:attention_level_congd}}
        
    \caption{Evaluation performance as $A$ changes for (a) EgoGesture, (b) IPN Hand, and (c) ConGD. For EgoGesture and ConGD MJI is used to evaluate performance. For IPN Hand we use LA.}
\end{figure*}

To determine the attention level parameter, $A$, we have evaluated the model on all three datasets with different values of $A$. Figure \ref{fig:attention_level_ego} illustrates the impact of the attention-level parameter on the MJI for the EgoGesture validation set and the highest value is obtained when $A=8$. Note that $A=1$ is simple concatenation of  features. For IPN hand dataset, we calculated the LA metric and achieved the highest result at $A=8$ (as shown in Figure \ref{fig:attention_level_ipn}). On the ConGD dataset (see Figure \ref{fig:attention_level_congd}) we obtained the highest MJI when $A=10$. Considering these findings we have conducted the remaining experiments considering the values of $A$ that achieved the highest MJI and LA results.      


\subsubsection{Impact of $\lambda_1$ and $\lambda_2$}

The impact of the proposed losses are controlled through the hyper-parameters, $\lambda_1$ and $\lambda_2$. In order to select the best values for the hyper-parameters we follow similar experiments to those in \cite{farha2019ms}. For the evaluations we have considered the EgoGesture dataset. 

The impact of $\lambda_1$ is measured by keeping the $\lambda_2$ fixed at $\lambda_2=0.25$. Then we train the model at different $\lambda_1$ values from 0.05 to 0.25. As shown in Table \ref{tab:hyper_para}, when $\lambda_1=0.15$ the model achieves the highest result, while there is a slight drop when $\lambda_1=0.05$ and $\lambda_1=0.25$. The reason for this is that the smoothing loss tends to heavily penalize changes in frame-wise labels, which controls the boundaries between action segments. 

To measure the impact of $\lambda_2$ we maintain a fixed $\lambda_1$ of 0.15 and trained models with different $\lambda_2$ values ranging from 0.05 to 0.35. As shown in the Table \ref{tab:hyper_para}, the highest result is obtained at $\lambda_2=0.25$. This indicates that our proposed mid-point based loss contributes more than the smoothing loss to the final gesture recognition result, and highlights the values of seeking to maintain natural gesture transitions. As $\lambda_2$ was increased further to $\lambda_2=0.35$, we noticed a slight drop in the overall result.  

Based on these findings, we use $\lambda_1=0.15$ and $\lambda_2=0.25$ for the remaining experiments.

\begin{table}[ht!]
\caption{Impact of $\lambda_1$ and $\lambda_2$ on the EgoGesture dataset with the window size of 16 and stride of 8.}
\centering
\resizebox{.75\linewidth}{!}{

\begin{tabular}{l|l}
\hline \hline
Imapct of $\lambda_1$          & MJI \\ \hline
TMMF $(\lambda_1=0.05, \lambda_2=0.25)$ &  0.786   \\ 
TMMF $(\lambda_1=0.15, \lambda_2=0.25)$ &  0.803   \\ 
TMMF $(\lambda_1=0.25, \lambda_2=0.25)$ &  0.791   \\ \hline 
Impact of $\lambda_2$           &  MJI   \\ \hline
TMMF $(\lambda_1=0.15, \lambda_2=0.05)$ &  0.742   \\ 
TMMF $(\lambda_1=0.15, \lambda_2=0.15)$ &  0.777   \\ 
TMMF $(\lambda_1=0.15, \lambda_2=0.25)$ &  0.803   \\ 
TMMF $(\lambda_1=0.15, \lambda_2=0.35)$ &  0.797   \\ \hline
\end{tabular}}
\label{tab:hyper_para}
\end{table}

\subsubsection{State-of-the-art Comparison}

In Table \ref{tab:egogesture1}, we compare the performance of the proposed method with the current state-of-the-art using the EgoGesture dataset. It should be noted that the comparison is done using continuous video streams containing multiple gestures in each videos (not pre-segmented videos containing a single gesture per video). For the evaluations, the MJI (as described in \ref{metrics}) is used. Similar to the original works \cite{zhang2018egogesture,cao2017egocentric}, we use two settings in evaluating the results. In both settings we keep the sliding window length at 16 and we consider strides of 16 (l=16, s=16) and 8 (l=16, s=8). In Table \ref{tab:egogesture1}, the method introduced in \cite{wang2016large} (denoted QOM + IDMM) is based on a two-stage paradigm which handles the temporal segmentation and classification tasks separately using only the depth information. In order to detect the start and the end frames of the each gesture, the Quantity of Movement (QOM) is used. Then each segmented gesture is passed through an Improved Depth Motion Map (IDMM) and fed to a ConvNet in order to perform gesture classification. 

In \cite{zhang2018egogesture}, the authors obtain the label predictions by considering the class probabilities of each clip predicted by the C3D softmax layer. Here, the sliding window is applied over the whole sequence to generate a video clip. For the setting where the sliding window overlaps (i.e. l=16, s=8), frame label predictions are obtained by accumulating the classification scores of two overlapped windows and the most likely class is chosen as the predicted label for each frame. The authors in \cite{zhang2018egogesture} further improved the result by utilising an LSTM network (denoted C3D + LSTM). In C3D + LSTM, the class labels of each frame are predicted by an LSTM based on the C3D features extracted at the current time slice. An LSTM with a hidden feature dimension of 256 is used. The authors have gained better results by employing the LSTM network. 

We also compare our results with the method introduced in \cite{cao2017egocentric} where the authors proposed a gesture prediction method which employed a Spatio-Temporal Transfer Module (denoted by C3D + STTM). However, in both settings, our proposed TMMF model is able to outperform the current state-of-the-art methods for the EgoGesture dataset by a considerable margin. We also obtained 1.9\% gain (metric calculation of MJI (see Sec. \ref{metrics})) using the second setting (i.e. l=16, s=8) where the sliding windows overlap.

In addition to the results included in the Table \ref{tab:egogesture1} in our original manuscript, we have evaluated the results using the frame-wise accuracy metric and achieved 95.10\%. Compared to the highest recorded state-of-the-art result of 94.72\% in \cite{shi2019gesture}, we achieved 0.38\% improvement in frame-wise accuracy. However, the limitation of frame-wise metrics is that they do not capture the segmentation behaviour of continuous data. Models achieving similar accuracy can have large variation in qualitative results \cite{lea2017temporal}. As such, we are unable to compare the segmentation abilities of our method with the method proposed in \cite{shi2019gesture}. 

\begin{table}[ht!]
\caption{Comparison of our proposed TMMF model with the state-of-the-art methods on the EgoGesture dataset. Results are shown using the MJI metric (see Sec. \ref{metrics}). }
\centering
\resizebox{.75\linewidth}{!}{
\begin{tabular}{ll}
\hline
\hline
Method              & MJI \\ \hline
QOM + IDMM \cite{wang2016large}    & 0.430   \\
C3D (l=16, s=16) \cite{zhang2018egogesture}     & 0.618   \\
C3D (l=16, s=8) \cite{zhang2018egogesture}      & 0.698   \\
C3D + STTM (l=16, s=8) \cite{cao2017egocentric} & 0.709   \\
C3D + LSTM (l=16, s=8) \cite{zhang2018egogesture}    & 0.718   \\ \hline
TMMF (l=16, s=16)   &    0.784     \\
TMMF (l=16, s=8)    &    \textbf{0.803}    \\ \hline
\end{tabular}}
\label{tab:egogesture1}
\end{table}

We also evaluate our proposed TMMF model on the IPN Hand dataset \cite{benitez2020ipn} and Table \ref{tab:ipnhands_1} includes a comparison of our proposed fusion model with the state-of-the-art. The results use the Levenshtein accuracy metric (see \ref{metrics}). In the original work \cite{benitez2020ipn}, the authors have performed continuous gesture recognition using a two-stage approach where at the first stage a separate detection model is used to detect gestures within a sequence. For this task binary classification is carried out to separate gestures from non-gestures using a ResLight-10 \cite{kopuklu2019real} model. In the second stage the detected gesture is classified by the classification model (ResNet50 or ResNetXt-101). For the overall process in \cite{benitez2020ipn}, the authors have considered different combinations of data modalities such as RGB-Flow and RGB-Seg where 'Flow' and 'Seg' refer to optical flow and semantic segmentation respectively. However, the authors gained the highest classification results using ResNetXt-101 with RGB-Flow data.

In contrast to the two-stage approach introduced in \cite{benitez2020ipn}, we use a single-stage method which directly predicts the sequence of gesture class labels for each frame of the entire sequence. Even though such a direct approach is challenging and requires a high level of discriminating ability within the model to separate multiple gesture and non-gesture classes, our fusion model outperforms the state-of-the-art results on IPN hand dataset by a significant margin. In Sec. \ref{scalability} we further evaluate the model using the three available modalities of RGB, flow and semantic segmentation outputs, illustrating the scalability of the proposed framework.

We further compare our proposed TMMF model with the state-of-the-arts on the ConGD dataset in Table \ref{tab:conGD_results}. The method introduced in \cite{pigou2017gesture} utilises deep residual networks fused with a bi-directional LSTM to learn spatio-temporal features, while in \cite{cihan2017particle} a probabilistic approach is used to segment gestures prior to the 3DCNN based recognition step. In \cite{liu2017continuous} a temporal segmentation method is proposed utilising the hand positions obtained through a fast-RCNN based network which is then followed by a classification step. A similar approach is taken in \cite{wang2017large} with a combination of a two-stream CNN based detection model followed by recognition models handling depth and RGB modalities. Compared to the previous methods, the models in \cite{zhu2018continuous,wan2020chalearn} are able to achieve the best results by a significant margin. \cite{zhu2018continuous} proposed a 2-stage method where the authors introduce a temporal dilated Res3D network for the gesture detection task, which is followed by a classification module based on a combination of 3DCNN, convolutional LSTM and 2D-CNNs. In \cite{wan2020chalearn}, a bi-directional LSTM is utilised to perform temporal segmentation task. Our proposed TMMF method outperforms the state-of-the-art models of \cite{zhu2018continuous,wan2020chalearn} by 1.38\% and 1.22\% respectively (results based on the MJI metric). 

\begin{table}[ht!]
\caption{Comparison of our proposed TMMF model with the state-of-the-art method on the IPN Hand dataset. The results are shown in term of Levenshtein accuracy (see Sec. \ref{metrics}). }
\centering
\resizebox{.8\linewidth}{!}{
\begin{tabular}{lll}
\hline
\hline
Method       & Modality & Results \\ \hline
ResNet50 \cite{benitez2020ipn}     & RGB-Seg  & 33.27   \\
ResNet50 \cite{benitez2020ipn}     & RGB-Flow & 39.47   \\
ResNeXt-101 \cite{benitez2020ipn} & RGB-Seg  & 39.01   \\
ResNeXt-101 \cite{benitez2020ipn} & RGB-Flow & 42.47   \\ \hline
TMMF & RGB-Flow &    \textbf{68.12}     \\ \hline
\end{tabular}}
\label{tab:ipnhands_1}
\end{table}

\begin{table}[ht!]
\caption{Comparison of our proposed TMMF model with the state-of-the-art methods on the LAP ConGD dataset. Results are shown using the MJI metric (see Sec. \ref{metrics}).}
\centering
\resizebox{.9\linewidth}{!}{
\begin{tabular}{ll}
\hline
\hline
Method              & MJI \\ \hline
Temporal Residual Networks \cite{pigou2017gesture}& 0.3164  \\
C3D + Probabilistic Forced Alignment \cite{cihan2017particle}  &  0.3744  \\
ConvNets + convLSTM \cite{wang2017large}     &  0.5950  \\
Faster RCNN + C3D \cite{liu2017continuous}&  0.6103  \\
TD-Res3D \cite{zhu2018continuous}   &  0.7163  \\ 
Bi-LSTM \cite{wan2020chalearn}   & 0.7179\\\hline
TMMF    &  \textbf{0.7301}      \\ \hline
\end{tabular}}
\label{tab:conGD_results}
\end{table}

\begin{figure*}[ht!]
    \centering
    \subfigure[][ConGD]{\includegraphics[width=.79\textwidth]{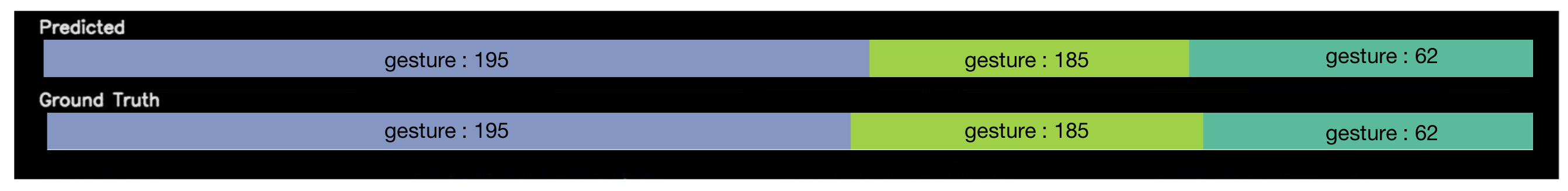}} \\
    \subfigure[][EgoGesture]{\includegraphics[width=.8\textwidth]{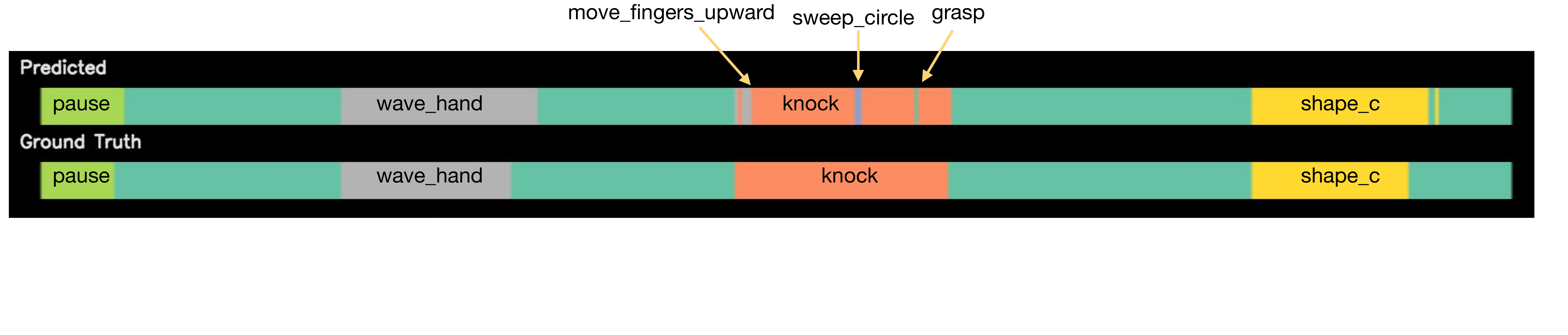}} \\
    \subfigure[][IPN Hand]{\includegraphics[width=.8 \textwidth]{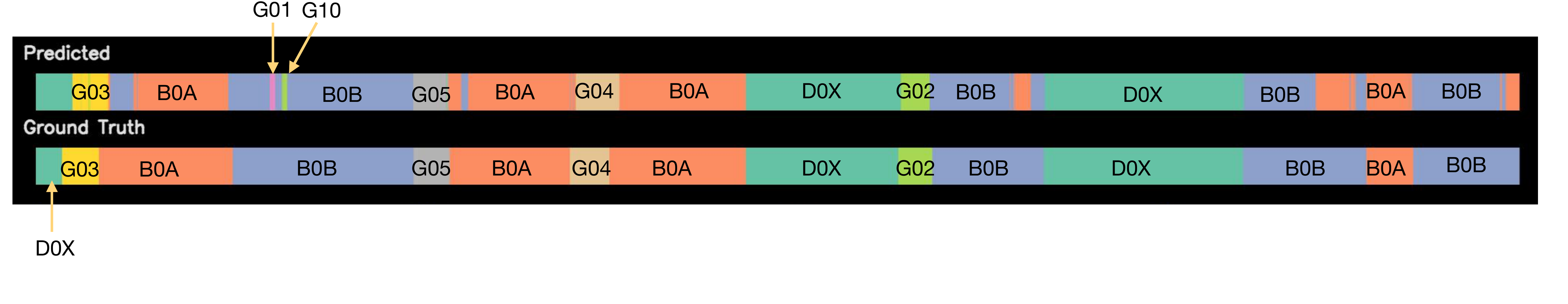}} \\
    \caption{Qualitative results of the propose model predictions on ConGD, EgoGesture, IPN Hand dataset.}
    \label{fig:qualitative1}
\end{figure*}

In addition to the quantitative results, we provide qualitative results (in Fig. \ref{fig:qualitative1} a,b, c) where we visualise the temporal gesture predictions generated by the proposed method for different frame sequences from ConGD, EgoGesture and IPN Hand datasets respectively. Please refer to supplementary material for additional qualitative results. 

\subsubsection{Impact of Loss Formulation}

We investigate the impact of our proposed loss formulation, which enhances the overall learning of the introduced model. In Table \ref{tab:loss_results}, the MJI of the EgoGesture dataset obtained with different loss formulations is shown. In the table, $\mathcal{L}_{ce}$, $\mathcal{L}_{sm}$ and $ \mathcal{L}_{med}$ represents the cross-entropy loss (Eq. \ref{eq:ce}), the smoothing loss (Eq. \ref{eq:sm}) and mid-point smoothing loss (Eq. \ref{eq:med}) respectively. Note that the combination of all three losses is the loss used by the proposed approach, $\mathcal{L}_{overall}$ (Eq. \ref{eq:overall}).   

\begin{table}[ht!]
\caption{The impact of different loss formulations: we compare the MJI on the EgoGesture dataset with different loss formulations. Here, $\mathcal{L}_{ce}$, $\mathcal{L}_{sm}$ and $ \mathcal{L}_{med}$ represent the cross-entropy loss (Eq. \ref{eq:ce}), the smoothing loss (Eq. \ref{eq:sm}) and mid-point smoothing loss (Eq. \ref{eq:med}) respectively.}
\centering
\resizebox{.55\linewidth}{!}{
\begin{tabular}{ll}
\hline
\hline
Loss & MJI \\ \hline
$\mathcal{L}_{ce}$    &     0.753    \\
$\mathcal{L}_{ce}+ \lambda_1 \mathcal{L}_{sm}$  &    0.781     \\
$\mathcal{L}_{ce}+ \lambda_2 \mathcal{L}_{mid}$ &    0.784     \\
$\mathcal{L}_{ce}+ \lambda_1 \mathcal{L}_{sm} + \lambda_2 \mathcal{L}_{mid}$  &   \textbf{0.803} \\\hline
\end{tabular}}
\label{tab:loss_results}
\end{table}

From Table \ref{tab:loss_results} we observe that both losses, $\mathcal{L}_{mid}$ and $\mathcal{L}_{sm}$ have contributed to improving the the cross-entropy loss, with the proposed mid-point based smoothing mechanism showing a slightly higher improvement. However, we observe a significant improvement when utilising all losses together, illustrating the importance of mid-point based comparison of different predicted and ground-truth windows. 

\subsubsection{Scalability of the Fusion Block}
\label{scalability}

In order to illustrate the scalability of the proposed framework and fusion mechanism to different numbers of modalities, we make use of a third modality: the segmentation maps which are provided in the IPN Hand dataset \cite{benitez2020ipn}. 

To make the feature extraction of hand segmentation maps more meaningful, we use the Pix-to-Pix GAN introduced in \cite{isola2017image} \footnote{We use the implementation provided at https://github.com/junyanz/pytorch-CycleGAN-and-pix2pix}. We first train the GAN to generate hand segmentation maps that are similar to the ones provided with the IPN hand dataset. We set the number of filters in the generator and the discriminator to 8 and train the GAN by following the original work. After GAN training, we use the trained generator model for feature extraction of the third modality where features are obtained from the bottleneck layer of the generator. These extracted feature vectors (of dimension $ 18 \times 18 \times 8 = 2592$ ) are fed to the third UFM model along side the UFM models used for the RGB and optical flow based feature vectors, as per the model evaluated in Tab. \ref{tab:ipnhands_1}. It should be noted that having varying feature vector dimensions (i.e 2592 for segmentation map inputs while 2048 for RGB and optical flow features) does not effect the fusion as the UFM block maps the feature vectors to the same dimensionality at the output head, which is the input to the fusion block. As expected, with the use of three modalities we were able to improve the overall Levenshtein accuracy by 1.8\% from the setting with only 2 modalities, achieving a Levenshtein accuracy of 69.92\% with the 3 feature modalities. With this evaluation we illustrate that the proposed method can seamlessly be extended to fuse the data from a different number of modalities with different feature dimensions. 

\subsubsection{Ablation Experiment}
\label{subsec:ablation_exp}

\begin{figure*}[ht!]
    \centering
    \subfigure[][Only RGB/ Only Depth]{\includegraphics[width=.25
    \linewidth]{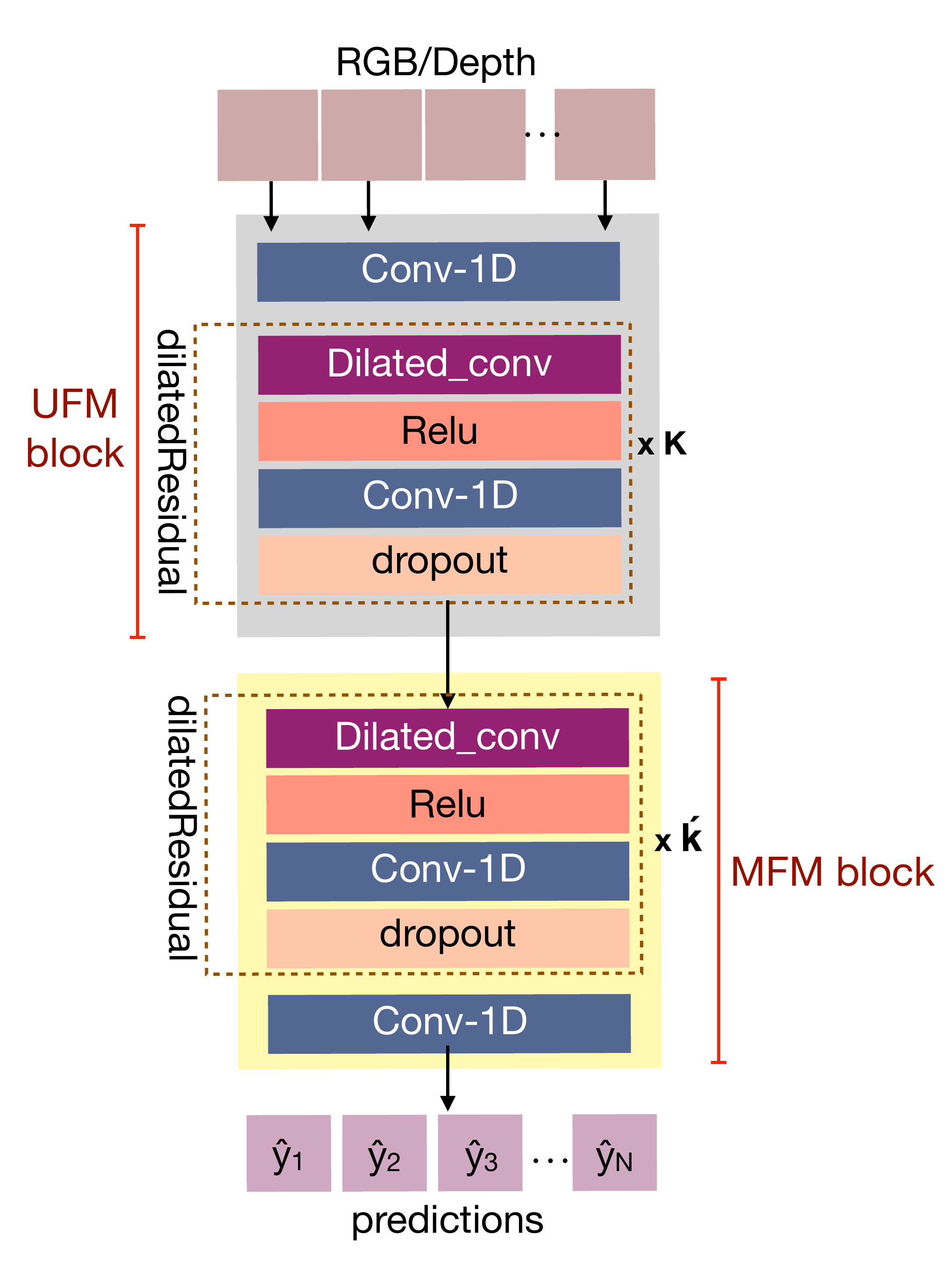}} 
    \subfigure[][Simple Fusion]{\includegraphics[width=.4\linewidth]{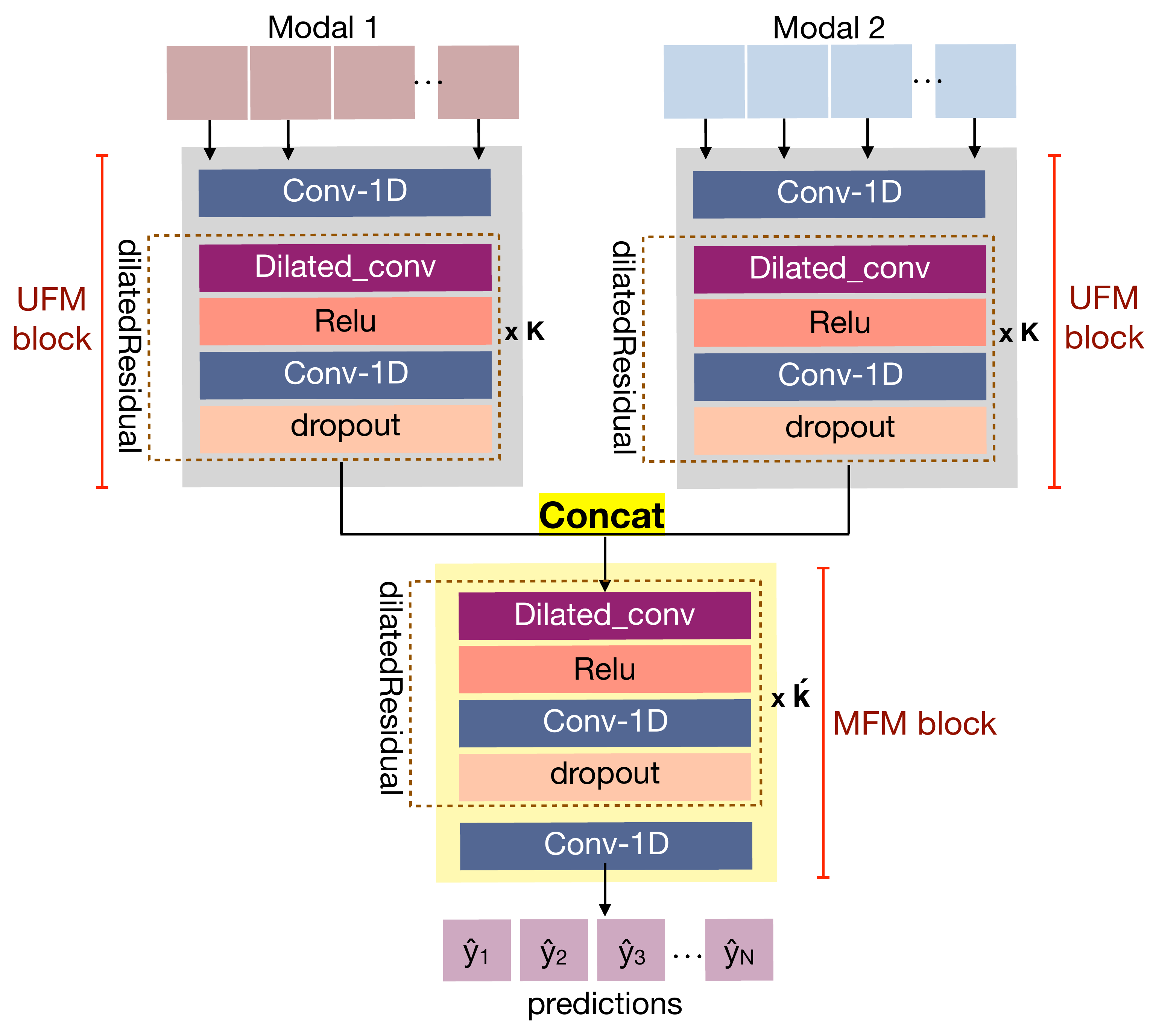}}
    \caption{Ablation models: (a) The models that utilise a single modality (Only RGB/ Only Depth) are composed of a single UFM and MFM block, and the MFM block works as a second uni-modal network. (b) The Simple Fusion model is formulated by performing concatenating the sequence of RGB and depth features instead of utilising the proposed fusion block.}
    \label{fig:ablation_arc}
\end{figure*}

In order to demonstrate the importance of the proposed fusion mechanism we conducted an ablation experiment. In the experiment, we gradually remove components of the proposed framework and re-train and test the models with the EgoGesture dataset. In Table \ref{tab:ablation}, we report the evaluation results for five ablation models with the MJI metric. The ablation models are formulated as follows.

\begin{itemize}
    \item \textbf{Only RGB:} A single UFM block is utilised for the RGB stream and the output of the uni-modal block is passed directly through the MFM block (see Fig. \ref{fig:ablation_arc} (a)). Here, the MFM block works as a second uni-modal network as only a single modality is used. Therefore, the proposed feature enhancer (FE) is not used.
    
    \item \textbf{Only Depth:} The model is similar to that of the 'Only RGB' stream ablation model ((see Fig. \ref{fig:ablation_arc} (a))). However, the instead of the input RGB stream, the depth input stream is used.
    
    \item \textbf{Simple Fusion:} Our proposed framework is utilised without the introduced fusion block. The fusion is performed by concatenating along the sequence of RGB and depth modalities. The model architecture is illustrated in Fig \ref{fig:ablation_arc} (b).
    
    \item \textbf{TMMF w/o UFM:} Our proposed framework without the UFM blocks to temporally map and encode the uni-modal features. The sequence of features extracted through the ResNet50 with dimension 2048 is fed directly to the fusion block.  
    
    \item \textbf{TMMF w/o MFM:} The proposed TMMF model without the MFM block to temporally map the multi-modal features and classify the gestures. The output form the fusion block is directly used for classification.  
    
    \item \textbf{TMMF w/o FE:} The proposed framework without the feature enhancer (FE) module. 
\end{itemize}

\begin{table}[ht!]
\caption{Evaluation results for the ablation models using the EgoGesture dataset.}
\centering
\resizebox{.55\linewidth}{!}{
\begin{tabular}{ll}
\hline
\hline
Model           & MJI \\ \hline
Only RGB        &    0.697     \\
Only Depth      &    0.741     \\
Simple Fusion   &    0.755    \\\hline
TMMF w/o UFM   &    0.640   \\
TMMF w/o MFM   &   0.763     \\
TMMF w/o FE &    0.792     \\ \hline
TMMF        &    \textbf{0.803}    \\ \hline
\end{tabular}}
\label{tab:ablation}
\end{table}

A key observation based on the results presented in \ref{tab:ablation} is that naive concatenation of multi-modal features does not generate helpful information for continuous gesture recognition. We observe a performance drop of approximately 5\% when simple concatenation is applied in comparison to the proposed approach, and only a slight improvement for naive concatenation over the best individual mode (depth). 

We also have conducted experiments by removing the important components of the proposed TMMF model. It is clear that the UFM block plays a major role at the earlier stage of the framework, by learning the spatio-temporal information of the uni-modal data and encode this into more discriminative semantic features to support the fusion bock. The results of `TMMF w/o UFM' show that the performance is even lower than the `only RGB' setting which utilises a UFM model for the uni-modal feature mapping. We performed another evaluation without the MFM block (TMMF w/o MFM) to map the multi-modal features while keeping the UFM blocks. The evaluation result is slightly higher than the `Simple Fusion' and we believe this mainly benefits from the UFM and fusion blocks. Furthermore, the proposed temporal fusion strategy, as well as the feature enhancement block (based on the results on the TMMF w/o FE setting), have clearly contributed to the superior results that we achieved. 

\section{Conclusion}
We propose a single-stage continuous gesture recognition framework (TMMF) with a novel fusion method to perform multi-modal feature fusion. The proposed framework can be applied to varying length gesture videos, and is able to perform the gesture detection and classification in a single direct step without the help of an additional detector model. The proposed fusion model is introduced to handle multiple modalities without a restriction on the number of modes, and further experiments demonstrate the scalability of the fusion method and show how the multiple streams complement the overall gesture recognition process. With the proposed loss formulation our introduced single-stage continuous gesture recognition framework learns the gesture transitions with considerable accuracy, even with the rapid gesture transitions of the IPN hand dataset. The ablation experiment further highlight the importance of the components of the proposed method, which outperformed the state-of-the-art systems on all three datasets by a significant margin. Our model has applications to multiple real-world domains that require  classification on continuous data, while the fusion model is applicable to other fusion problems where videos or signal inputs are present, and can be used with or without the UFM or MFM blocks. 



\section*{Acknowledgment}

The research presented in this paper was supported by an Australian Research Council (ARC) Discovery grant DP170100632.

\ifCLASSOPTIONcaptionsoff
  \newpage
\fi



\bibliographystyle{IEEEtran}
\bibliography{IEEEexample.bib}

%





\end{document}